\documentclass[sigconf]{acmart}
\usepackage{booktabs}
\usepackage{multirow}
\usepackage{geometry}
\usepackage{graphicx}
\usepackage{makecell} 
\usepackage{adjustbox}
\usepackage{subcaption}
\usepackage{enumitem}
\usepackage{xcolor}
\usepackage{arydshln}     
\usepackage{siunitx}      

\AtBeginDocument{%
  }

\setcopyright{acmlicensed}
\copyrightyear{2025}
\acmYear{2025}
\acmDOI{XXXXXXX.XXXXXXX}
\acmConference[Conference acronym 'XX]{Make sure to enter the correct
  conference title from your rights confirmation email}{June 03--05,
  2018}{Woodstock, NY}
\acmISBN{978-1-4503-XXXX-X/2018/06}

\begin{document}

\title{OneVision: An End-to-End Generative Framework for \\
Multi-view E-commerce Vision Search}

\author{Zexin Zheng$^*$, Huangyu Dai$^*$, Lingtao Mao$^*$, Xinyu Sun, Zihan Liang, \\ Ben Chen$^{\dagger}$, Yuqing Ding, Chenyi Lei$^{\dagger}$, 
Wenwu Ou, Han Li, Kun Gai}
\thanks{*Equal Contribution.}
\thanks{$\dagger$ Corresponding author.}
\affiliation{%
  \institution{Kuaishou Technology}
  \city{Beijing}
  \country{China}
}

\renewcommand{\shortauthors}{Trovato et al.}

\begin{abstract}
Traditional vision search, similar to search and recommendation systems, follows the multi-stage cascading architecture (MCA) paradigm to balance efficiency and conversion. Specifically, the query image undergoes feature extraction, recall, pre-ranking, and ranking stages, ultimately presenting the user with semantically similar products that meet their preferences. This multi-view representation discrepancy of the same object in the query and the optimization objective collide across these stages, making it difficult to achieve Pareto optimality in both user experience and conversion. In this paper, an end-to-end generative framework, \textbf{OneVision}, is proposed to address these problems. OneVision builds on VRQ, a vision-aligned residual quantization encoding, 
which can align the vastly different representations of an object across multiple viewpoints while preserving the distinctive features of each product as much as possible. Then a multi-stage semantic alignment scheme is adopted to maintain strong visual similarity priors while effectively incorporating user-specific information for personalized preference generation.
In offline evaluations, OneVision performs on par with online MCA, while improving inference efficiency by 21\% through dynamic pruning; In A/B tests, it achieves significant online improvements: +2.15\% item CTR, +2.27\% CVR, and +3.12\% order volume. These results demonstrate that a semantic ID centric, generative architecture can unify retrieval and personalization while simplifying the serving pathway.

\end{abstract}


\maketitle

\section{Introduction}

\begin{figure}
  \includegraphics[width=0.475\textwidth]{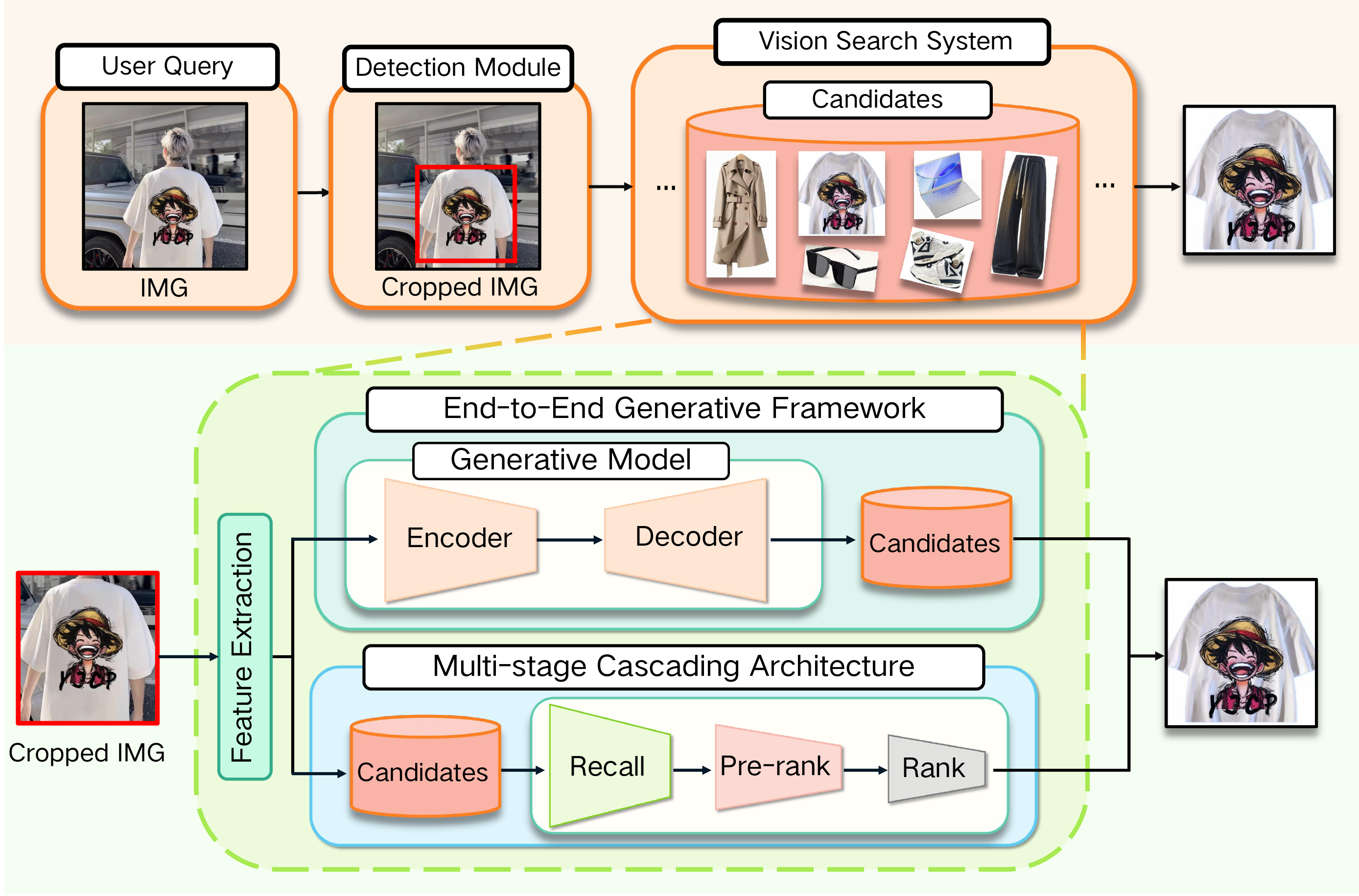}
  \centering
  \caption{Comparison between the proposed end-to-end generative framework and the traditional multi-stage cascading architecture in e-commerce vision search.}
  \label{fig1}
  \vspace{-3mm}
\end{figure}

A vision search system takes an image as the input query and retrieves results with high visual relevance. Similar to other search engines, such systems are typically organized as a multi-stage pipeline, as depicted in Figure \ref{fig1}:
(i) feature extraction, which obtains comprehensive visual representations from regions of interest;
(ii) the recall stage, which constructs a large pool of relevant candidates by retrieving items with high visual similarity;
(iii) the pre-ranking stage, which further filters and refines the candidate pool; and
(iv) the ranking stage, which orders the remaining candidates to deliver personalized retrieval results. Unlike text search systems, where semantic relevance and user personalization can often be optimized jointly from the start, vision search must first ensure that the recalled candidates are highly visually similar before user-specific signals can be effectively applied at the ranking stage.

Recent years have witnessed substantial progress in vision search, propelled by advances across multiple fronts, including object detection~\cite{cheng2024yolow,tian2025yolov12,wang2025yoloe}, category prediction~\cite{zhu2022enhanced,noor2023bottom,noor2024consistency}, and visual representation learning~\cite{lee2022correlation,shao2023global,liang2025uniecs,zhai2019learning,simeoni2025dinov3}. These developments have driven the construction of large-scale industrial systems, with major e-commerce platforms such as eBay~\cite{yang2017visual} and Alibaba~\cite{zhang2018visual, chen2024branches} deploying comprehensive vision search pipelines to support production-scale applications. Despite these advances, traditional pipelines in vision search still face structural limitations: (i) Difficulty in handling multi-view discrepancy. Query images and product images often differ significantly in viewpoint, making it hard for conventional retrieval pipelines to align their visual representations and recall the correct products. (ii) Stage-wise objective misalignment. The optimization goals across three stages collide, making it difficult to reach Pareto optimality in both user experience and conversion. (iii) Heavy multi-stage overhead. The multi-stage architecture introduces substantial computational cost and cross-module communication, increasing latency and complicating system maintenance.

Over the past two years, generative retrieval (GR) has emerged as a new paradigm for large-scale search, reframing retrieval as a sequence-to-sequence generation. Instead of relying on multi-stage matching, GR models directly generate discrete item identifiers from a query and user-behaviour sequence, thereby alleviating the limitations of traditional retrieval architectures. OneRec~\cite{deng2025onerec} delivers a single-stage industrial recommender that unifies recall, pre-ranking, and ranking in one model. In e-commerce search, OneSearch~\cite{chen2025onesearch} provides the first end-to-end generative retrieval framework with superior performance for high-quality recall and ranking. For image retrieval specifically, IRGen~\cite{zhang2024irgen} formulates image retrieval as seq2seq generation of nearest-neighbor IDs. GENIUS~\cite{kim2025genius} adopts modality-decoupled semantic quantization and a query augmentation to enhance generalization. However, IRGen and GENIUS overlook multi-view discrepancies of the same object and ignore user-specific information, limiting their effectiveness in industrial vision search applications. 

To address the limitations of traditional MCA pipelines and the issues neglected by current GR, in this paper, we propose \textbf{OneVision}, an end-to-end vision search system for the e-commerce scenario that effectively integrates the feature extraction, recall, pre-ranking, and ranking stages. Our framework effectively captures multi-view representations of the same object in e-commerce scenarios and incorporates user-specific behaviours, ensuring that the generated candidates are both visually relevant and user-preferred.
Specifically, Onevision includes the following components:

1) \textbf{Vision-aligned residual quantization (VRQ) encoding. } 
VRQ aligns the diverse visual representations of the same object across multiple viewpoints while preserving product-specific discriminative features for residual encoding. A multi-view contrastive objective guides residual quantization in the shallow codebook layers. At deeper layers, VRQ retains fine-grained residuals and integrates statistical business attributes, and OPQ is applied to jointly quantize these item-specific residual features. Category information is incorporated to enforce category–code consistency. This design achieves coarse semantic alignment while preserving the fine-grained visual distinctions for accurate product representation.

2) \textbf{Multi-stage pipeline and personalized generation. } 
A multi-stage pipeline for generative training with user behavior enables the model to generate candidates that are both visually similar and well aligned with individual user preferences. The training process consists of three stages: Pretraining, Supervised Fine-Tuning (SFT), and Direct Preference Optimization (DPO).
Pretraining is used for initial semantic alignment among product images, textual information, and semantic IDs (SID).
SFT facilitates multi-view collaborative feature learning to better capture the semantic relationships between the query image and candidate items. Finally, user behavioral sequences are constructed to model preferences, and list-wise DPO is applied for personalized generation.

3) \textbf{Dynamic pruning for efficient inference. } 
To address redundancy and inefficiency in image tokens, dynamic pruning preserves only the most semantically informative visual tokens.
Specifically, tokens are compressed before the decoder using K-means++ clustering, and potential performance degradation is mitigated via curriculum learning and distillation. This accelerates inference while maintaining accuracy and lowering compute cost.

\begin{figure}
\includegraphics[width=0.475\textwidth]{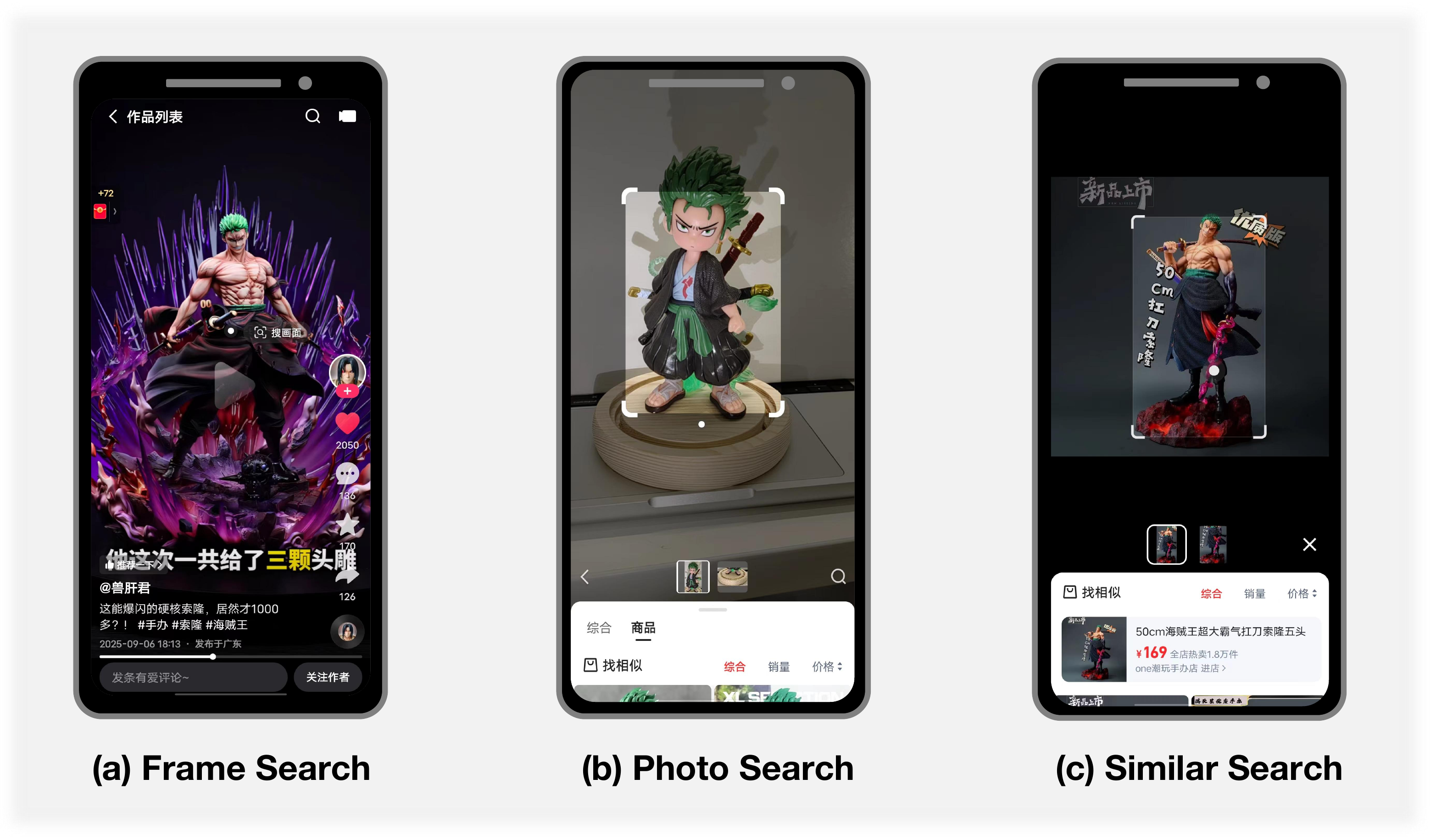}
  \centering
  \vspace{-5mm}  
  \caption{The major vision search scenarios in e-commerce on the Kuaishou Platform.}
  \label{fig2}
  \vspace{-2.8mm}
\end{figure}

Extensive offline evaluations and ablation studies show that OneVision matches the performance of the online MCA, while dynamic pruning boosts inference latency by 21\%. Large-scale online A/B tests on Kuaishou’s search platform further reveal notable improvements: CTR +2.15\%, CVR +2.27\%, OPM +3.60\%, Clicks +1.80\%, and Orders +3.12\%.
It demonstrates that OneVision effectively unifies retrieval and personalization, delivering measurable gains while simplifying the traditional serving pipeline.

\section{Related Works}
\subsection{Vision Search System}

Modern image-based search systems have advanced considerably with progress across core computer vision tasks such as object detection~\cite{cheng2024yolow,tian2025yolov12,wang2025yoloe}, category prediction~\cite{zhu2022enhanced,noor2023bottom,noor2024consistency}, and representation learning~\cite{lee2022correlation,shao2023global,liang2025uniecs,zhai2019learning,simeoni2025dinov3}. These advances have substantially improved feature extraction, enabling the generation of compact and discriminative visual embeddings from regions of interest. Building on such features, large-scale approximate nearest neighbor (ANN) retrieval is used to construct candidate pools, which are subsequently refined and personalized through pre-ranking and ranking~\cite{dagan2021image,nadkarni2022visually,dagan2023shop,you2024can}.

Prominent industrial systems demonstrate how these techniques are deployed at scale. eBay leverages deep region-based feature extraction with large-scale ANN retrieval to support item discovery in production~\cite{yang2017visual}; Alibaba integrates hierarchical category-aware embeddings with cascaded ranking for personalized search and recommendation~\cite{zhang2018visual}; JD.com builds a real-time visual search system with distributed indexing and sub-second image updates to handle large e-commerce catalogs~\cite{li2018design}; Pinterest documents the evolution of its “Shop The Look” platform, iteratively improving detection, retrieval, and labeling to scale visual shopping~\cite{shiau2020shop}.

\begin{figure*}
  \includegraphics[width=0.955\textwidth]{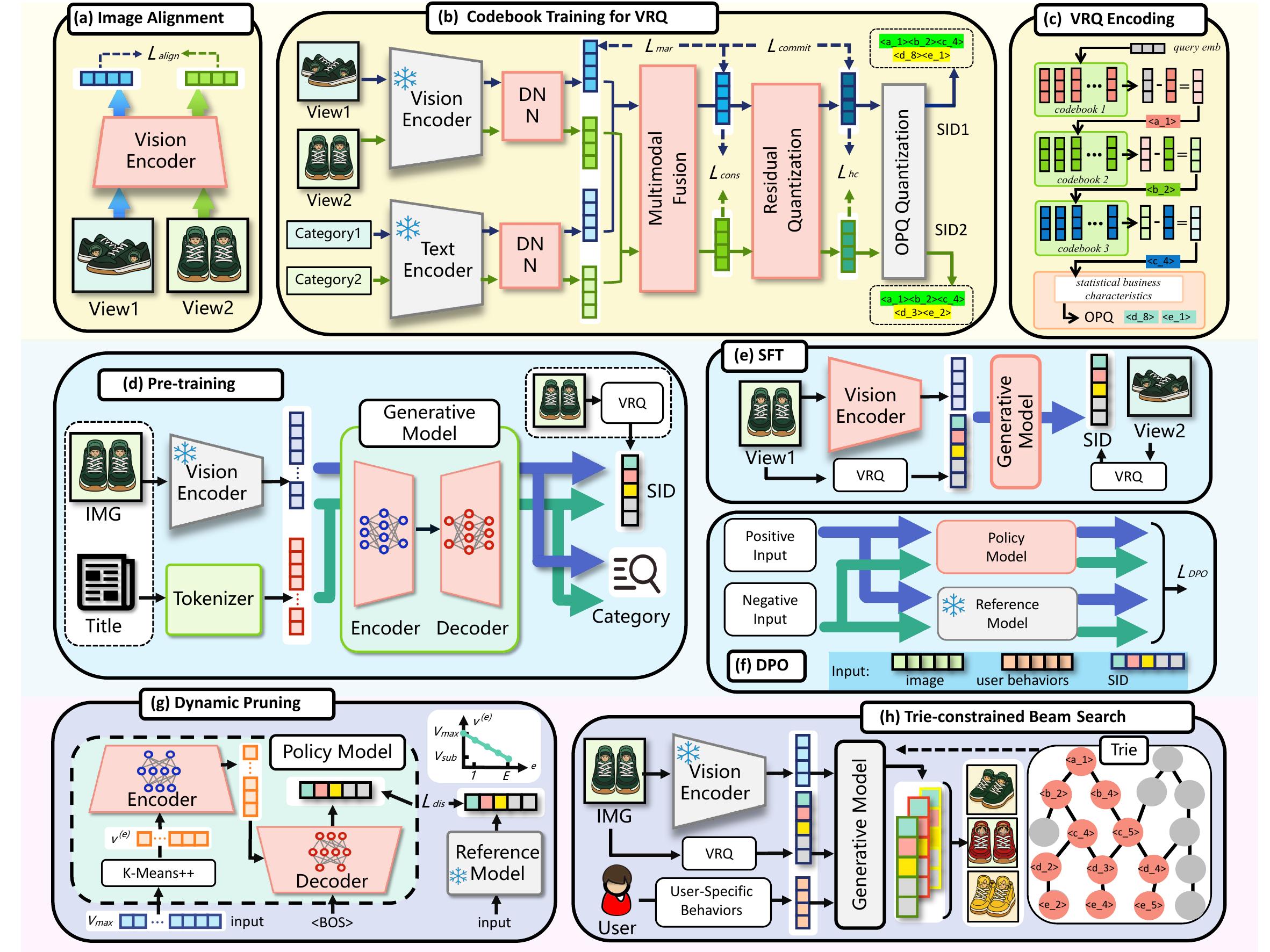}
  \centering
  \caption{Overall Framework of OneVision.
  The first row illustrates our VRQ codebook training and encoding process.
  The second row shows our multi-stage generative training pipeline, including Pretraining, SFT, and DPO.
The third row presents our dynamic pruning and end-to-end inference process.
  }
  \label{fig3}
\end{figure*}

\subsection{Generative Retrieval System}
\textbf{Generative Retrieval.} 
Generative Retrieval (GR) has recently gained wide attention for reframing large-scale retrieval as sequence-to-sequence generation. Instead of querying an external index with ANN, GR trains models to generate item identifiers (e.g., SIDs), outperforming multi-stage baselines like EBR \cite{huang2020embedding} and RocketQA \cite{qu2020rocketqa}. Notable advances include Tiger \cite{rajput2023recommender}, which derives semantic IDs from item content for end-to-end recommendation; DSI \cite{tay2022transformer}, which generates learned identifiers for direct retrieval; and LC-REC \cite{zheng2024adapting}, which adapts LLMs by integrating collaborative semantics through specialized tuning tasks.
\\
\textbf{Encode method in Generative Retrieval.}
Encoding is a core component of generative retrieval. 
RQ-VAE\cite{lee2022autoregressive} encodes images into residual-quantized codes, achieving high capacity with compact codebooks and modeling long-range dependencies efficiently via an autoregressive RQ-Transformer.
QARM (RQ-KMeans)\cite{luo2024qarm} aligns multimodal representations and compresses them into learnable SIDs through residual K-means, supporting end-to-end training and consistent semantics at scale. 
FSQ \cite{mentzer2023finite} replaces vector quantization with bounded scalar quantizers. This approach maintains performance while enabling large, well-utilized codebooks and easier optimization, which is particularly useful when reconstruction quality and sampling efficiency are critical.
\\
\textbf{Generative Retrieval application.} 
Recent advances show that GR models can replace traditional recall–ranking stacks. OneRec unifies recall and ranking via session-wise generation and DPO, achieving stable online gains with simpler serving~\cite{deng2025onerec}. MMQ further learns semantic IDs from multimodal content through a shared–specific tokenizer and behavior-aware fine-tuning to align semantics with user preferences~\cite{xu2025mmq}. In e-commerce search, OneSearch builds an end-to-end system where a keyword-enhanced hierarchical quantization encoder generates semantic IDs combining content, context, and collaborative signals; multi-view user sequences further improve personalization~\cite{chen2025onesearch}. These advances show GR can unify fragmented objectives, introduce personalization early, and reduce reliance on hand-crafted indices.
\\
\textbf{Generative Retrieval in image retrieval.} 
Recent work has adapted GR to image search by generating discrete identifiers for target items. IRGen \cite{zhang2024irgen} formulates retrieval as sequence-to-sequence generation of nearest-neighbor IDs with a semantic image tokenizer, enabling end-to-end training at million-item scale and near–real-time inference. GENIUS \cite{kim2025genius} further introduces modality-decoupled semantic quantization, employing a modality token followed by contrastively learned residual codes and incorporating query augmentation to improve generalization. Collectively, these methods replace index probing with code generation while maintaining scalability and efficiency, providing the basis for our image-centric, code-based vision search framework.

\section{Methodology}

In this section, we introduce OneVision, an end-to-end framework for multi-view e-commerce vision search. We first elaborate on the image hierarchical quantization encoding in~\S~\ref{sec:3.1}, and then detail the multi-stage pipeline for generative training in~\S~\ref{sec:3.2}. In~\S~\ref{sec:3.3}, we propose dynamic pruning for efficient online inference, and finally, in~\S~\ref{sec:3.4}, we outline the overall inference procedure. The complete pipeline is illustrated in Figure \ref{fig3}.
\subsection{Vision-aligned Residual Quantization}
\label{sec:3.1}
Semantic encoding of images is a key step for generative vision search. A common approach is hierarchical semantic encoding from coarse to fine levels. However, effective SIDs must satisfy two requirements: consistency, where similar items or multi-view images of the same object share shallow-level codes; and uniqueness, where different images remain distinguishable at deeper levels. 
However, current quantization methods face three major challenges:
(i) Techniques like FSQ~\cite{mentzer2023finite} and OPQ~\cite{ge2013optimized} struggle to hierarchically capture core attributes among similar items, limiting generative training;
(ii) General methods such as VQ-VAE~\cite{van2017neural}, RQ-VAE~\cite{lee2022autoregressive}, and RQ-KMeans~\cite{luo2024qarm} use shared codebooks, limiting their ability to preserve uniqueness;
(iii) Offline image encoding for retrieval often depends on object detection for item encoding, but incorrect main-subject recognition can yield erroneous embeddings and introduce noise.

A recent approach, RQ-OPQ~\cite{chen2025onesearch}, adopts RQ-Kmeans for shallow shared representations, and OPQ for deep unique features, which can encode each item hierarchically
and personally. Inspired by this schema, here we propose VRQ, which integrates multi-view contrastive learning for trainable shallow codebooks and leverages statistical business features in residual OPQ encoding to enhance consistency and uniqueness.
Additionally, we incorporate category information to suppress noisy samples and improve category-code consistency across the e-commerce image pool.

\subsubsection{Collaborative Image Feature Alignment}
In generative vision search systems, aligning similar product images or multi-view images enables the generation of high-quality visual representations, which are crucial for semantic encoding. 
Firstly, we construct a high-quality dataset of image pairs derived from real e-commerce user search logs, including multi-view images of the same items, query–item, and item–item pairs.
Then, we apply cosine similarity for preliminary filtering, employ Qwen-VL~\cite{bai2023qwen} to further verify semantic consistency, and finally perform manual inspection and refinement to remove noisy data.

We then apply a joint training objective to train the backbone $\mathcal{V}$ and align image features. A contrastive loss $\mathcal{L}_{\mathrm{cl}}$~\cite{chen2020simple} pulls positive pairs closer, while a circle loss $\mathcal{L}_{\mathrm{circle}}$~\cite{sun2020circle} improves separation between positives and negatives.
The total loss is:
\begin{equation}
\label{eq:1}
\mathcal{L}_{\mathrm{align}} = \lambda_{1}\,\mathcal{L}_{\mathrm{cl}} + \lambda_{2}\,\mathcal{L}_{\mathrm{circle}},
\end{equation}
where $\lambda_{1}$ and $\lambda_{2}$ are adjustable parameters.

\subsubsection{Metrics for Vision Quantization Encoding}
With high-quality image representations, the subsequent focus is on designing a semantic encoding scheme for product images to ensure both consistency and uniqueness. Accordingly, we propose Quantized ANN Score (QAS) and Independent Code Occupancy (ICO) as metrics for evaluating the quality of image semantic encoding. QAS quantifies how well semantically similar products stay clustered after quantization via ANN retrieval, while ICO measures the average number of items per unique SID. A higher QAS and a lower ICO, indicate stronger encoding consistency and uniqueness.

As shown in Table \ref{tab:corr_ablation}, current hierarchical encoding methods, including RQ-VAE and RQ-KMeans, are found to exhibit evident limitations on large and complex e-commerce image corpora; a detailed analysis is provided in~\S~\ref{sec:4.3}.
To enhance encoding quality, we integrate multi-view contrastive learning and a hierarchical consistency loss into shallow-level codebook training.
Furthermore, we incorporate category information to reduce noisy SIDs caused by main-subject misrecognition from inaccurate detection boxes.

\subsubsection{VRQ Hierarchical Quantization Encoding}
Concretely, given a single product image and its associated category label, we first extract the visual feature $x$ using the image encoder $\mathcal{V}$ and the category feature $y$ using a pretrained BGE model~\cite{xiao2024c}.
We then learn latent representations $v:=\mathcal{E}_v(x)$ and $t:=\mathcal{E}_t(y)$, where $\mathcal{E}_v$ and $\mathcal{E}_t$ are modality-specific DNN encoders for the visual and textual inputs, respectively.
To effectively combine multi-modal features into a unified representation~\cite{baldrati2023composed}, we compute the fused embedding as
\begin{equation}
\label{eq:2}
f=(1-\alpha) \cdot v+ \alpha\cdot t+f_{cat},
\end{equation}
where $f_{cat}$ is a combined feature derived from the concatenated image–text embedding through an MLP. 
The fusion weight $\alpha$ is dynamically learned from the concatenated features via another MLP followed by a sigmoid activation.

Additionally, inspired by~\cite{liang2025uniecs}, our multi-view contrastive objective comprises two key components: fused feature consistency and cross-modal alignment.  
The loss encourages the representations of paired, semantically similar products to remain close in the embedding space. Formally:
\begin{equation}
\begin{split}
\mathcal{L}_{\mathrm{cons}}
= -\frac{1}{N}\sum_{i=1}^N \big[
    \log \frac{\exp\big(f_i^{(1)} \cdot f_i^{(2)} / \tau\big)}
              {\sum_{j=1}^N \exp\big(f_i^{(1)} \cdot f_j^{(2)} / \tau\big)} \\
    + \log \frac{\exp\big(v_i^{(1)} \cdot f_i^{(2)} / \tau\big)}
              {\sum_{j=1}^N \exp\big(v_i^{(1)} \cdot f_j^{(2)} / \tau\big)}
\big],
\end{split}
\label{eq:3}
\end{equation}
where $f_i^{(1)}$ and $f_i^{(2)}$ denote fused embeddings of the $i$-th positive item pair, $N$ is the number of positive pairs in a batch, and $\tau$ is a temperature parameter.
Using only $\mathcal{L}_{\text{cons}}$ tends to make the model focus solely on visual features. To further enhance the category information~\cite{chen2023unified}, we introduce a margin loss:
\begin{equation}
\mathcal{L}_{\mathrm{mar}}
= \frac{1}{N} \sum_{i=1}^{N} \sum_{j=1}^{N}
\left[ \max\bigl(0,\ -\gamma + s_{ij}^{f2f} - s_{ij}^{v2f}\bigr) \right],
\label{eq:4}
\end{equation}
where $s_{ij}^{f2f}$ and $s_{ij}^{v2f}$ respectively denote the similarity between fused representations and the similarity between visual and fused representations, and $\gamma$ is a margin parameter.

Then the fused embedding $f$ is quantized by a $L$-level RQ-VAE and the codebook at level $l$ is $E_{l} = \{ e_{k}^{(l)} \in \mathbb{R}^{d} \,|\, k = 1,\dots,K_{l}\}$.
Specifically, the code $c_l$ at level $l$ is determined from the residual $r_l$ ($r_0=f$) as 
\begin{equation}
c_l = \arg\min_{k} \| r_l - e_k^{(l)} \| ,
\label{eq:5}
\end{equation}
and the residual is updated as $r_{l+1}=r_l-e_{c_l}^{(l)}$. Finally, we can obtain the SID ($c_0,...,c_{L-1}$) and the quantized representation at level $l$ is $\hat{f}^{(l)} = \sum_{i=0}^{l} e_{c_i}^{(i)} .
$
The commitment loss in RQ-VAE~\cite{lee2022autoregressive} is
\begin{equation}
\mathcal{L}_{\mathrm{commit}} = \sum_{l=0}^{L-1} \| r_{l} - \mathrm{sg}(e_{c_l}^{(l)} ) \|^{2},
\label{eq:5}
\end{equation}
where sg($\cdot$) is the stop-gradient operator and codebooks are updated using an exponential moving average (EMA)~\cite{razavi2019generating} to stabilize training.
Furthermore, we incorporate a hierarchical consistency loss to enforce encoding consistency for multi-view paired samples:
\begin{equation}
\mathcal{L}_{\mathrm{hc}} = \sum_{l=0}^{L-1} \| \hat{f}_i^{(l)} - \hat{f}_j^{(l)} \|^{2},
\label{eq:6}
\end{equation}
where $ \hat{f}_i^{(l)}$ and $ \hat{f}_j^{(l)}$ are the quantized paired vectors, and the overall training loss is the combination of all the aforementioned losses:
\begin{equation}
\mathcal{L}_{\mathrm{rq}} =\beta_1 \mathcal{L}_{\mathrm{cons}}+\beta_2\mathcal{L}_{\mathrm{mar}}+\beta_3\mathcal{L}_{\mathrm{commit}}+\beta_4\mathcal{L}_{\mathrm{hc}},
\label{eq:7}
\end{equation}
where $\beta_i$ are the adjustable parameters.
Notably, to achieve more robust RQ-VAE codebook training, we perform RQ-KMeans first and then adopt the resulting cluster centroids as the initialization for the RQ-VAE codebooks.

To further ensure the uniqueness of SIDs, we retain the residual embeddings produced by RQ-VAE, which capture fine-grained visual attributes. 
Meanwhile, to better distinguish visually identical products in e-commerce, we incorporate each item’s statistical business characteristics, such as the number of clicks, price, and purchase counts within a certain period. 
We then fuse these features with the RQ-VAE residuals and apply OPQ for residual encoding. 
Our VRQ encoding achieves significant improvements across multiple evaluation metrics, as detailed in~\S~\ref{sec:4.2} and ~\S~\ref{sec:4.3}.

\subsection{Multi-stage Generative Pipeline}
\label{sec:3.2}
In this section, we introduce the OneVision framework and its multi-stage learning. 
Traditional retrieval pipelines struggle with multi-view discrepancies between query and product images.
In contrast, generative models capture consistent multi-view representations and integrate user-behavior signals, producing candidates that are visually relevant and personalized.
Specifically, OneVision can be built using either encoder–decoder architectures (e.g., BART~\cite{lewis2019bart}, mT5~\cite{xue2020mt5}) or decoder-only backbones (e.g., Qwen3~\cite{yang2025qwen3}). For practical deployment, we adopt the encoder–decoder variant to speed up training and inference, as illustrated in Figure \ref{fig3}.

\subsubsection{Pretraining for Visual–Semantic Alignment}
Considering that the basic architectures, such as BART and mT5, are pretrained on large-scale textual corpora, while the input of the vision search system is based on image representations, we first introduce a pretraining stage to establish initial semantic alignment among product images, titles, categories, and SIDs.
To this end, we design four related tasks for pretraining: 
(i) Predicting the product SID from the corresponding image. 
(ii) Input both image and product title and output the relevant SID. 
(iii) Predicting the product category from the image. 
(iv) Input both image and product title and output the corresponding category.
The first task focuses on aligning visual representations with SIDs, while the remaining tasks introduce additional modalities to reinforce semantic relevance and facilitate alignment.
During unified training, we insert a start token $t_{[BOS]}$ and an ending token $t_{[EOS]}$ at the first and last place, and use a separate token $t_{[SEP]}$ between different modalities.
The training loss for the generative model is a cross-entropy loss for next-token prediction, formulated
as:
\begin{equation}
\mathcal{L}_{\mathrm{NTP}}
= -\sum_{k=1}^{T} \log P\bigl(t_k \mid u,t_{<k}\bigr).
\label{eq:8}
\end{equation}
where $t_k$ denotes the predicted next token, $t_{<k}$ represents the sequence of preceding tokens, $u$ is the input, and $T$ is the sequence length.
Specifically, for SID prediction, $T$ is set to the fixed length $L$.
\subsubsection{Supervised Fine-tuning for Collaborative Feature Learning}
For generative vision search, the goal is not only to align the images with their corresponding SIDs, but also to directly generate relevant candidate items from the query image. 
In this process, images of the same product taken from different views or angles must be mapped to the SID of the matching product. This allows the model to capture the intrinsic semantic and collaborative relations between query images and candidate items.

Specifically, after the pretraining phase, we introduce an SFT stage for collaborative feature learning.
This stage takes the query image and its corresponding SID as input and outputs the SID of the matched product.
Robust semantic understanding provides the foundation for incorporating user behaviour signals and enabling personalized ranking.
 
\subsubsection{Personalized Modeling and Post-training}
To model user behaviour, we incorporate two complementary, per-user components: (i) user long-term behavioural sequence, represented as a single aggregated SID obtained by fusing the SIDs of multiple products that the user has recently clicked or purchased on the Kuaishou e-commerce platform; and (ii) user short-term behavioural sequence, comprising the individual SIDs of the user’s five most recently clicked items during image-based search. In addition, we introduce explicit search-scene information to better distinguish intent across different contexts. The model is further trained with personalized SFT using the NTP loss to enhance its ability to capture nuanced user-specific purchasing preferences.

To further align with real user preferences, we adopt a session-based optimization strategy. User interactions within a session are categorized into three types: purchased items, clicked-but-not-purchased items, and exposed-but-not-clicked items. 
Purchased items are treated as positive samples, while the others serve as negative samples. 
A list-wise DPO objective is then applied to encourage higher scores for positive items across the entire set of candidates~\cite{chen2024softmax}.
Formally, for a given session $s$, with context $x_s$, let the candidate set be
$\mathcal{Y}_s = \{y^{+}, y_1^{-}, y_2^{-}, \dots, y_m^{-}\}$,
where $y^{+}$ denotes the purchased item and $y_j^{-}$ denotes negative items. The DPO objective is defined as:
\vspace{-0.5em}
\begin{equation}
\begin{aligned}
\mathcal{L}_{\mathrm{DPO}}
&= -\,\mathbb{E}\Big[
\log \sigma\Big(
-\log \sum_{y^{-} \in \mathcal{Y}^{-}}
\exp\!\Big(
\beta \log
\frac{\pi_{\theta}(y^{-}\!\mid x_s)}{\pi_{\mathrm{ref}}(y^{-}\!\mid x_s)}
\\[-2pt]
&\qquad
-\,
\beta \log
\frac{\pi_{\theta}(y^{+}\!\mid x_s)}{\pi_{\mathrm{ref}}(y^{+}\!\mid x_s)}
\Big)\Big)\Big],
\end{aligned}
\label{eq:list_dpo}
\end{equation}
where $\mathcal{Y}^-$ is the negative set, $\beta$ is the inverse temperature parameter, $\pi_{\mathrm{\theta}}$ is the policy model and $\pi_{\mathrm{ref}}$ is the reference model.

\subsection{Dynamic Pruning}
\label{sec:3.3}
While the GR model offers strong semantic understanding, it also introduces significant computational overhead—particularly during beam search decoding.
For a single query image, the model must perform $B$ rounds of decoding, where $B$ is the number of beam search candidates. This greatly impacts online inference speed.
Moreover, the encoding of image tokens contains redundancy and inefficiency, both in the visual encoder and the generative model’s encoder.
To mitigate this, we propose a dynamic pruning strategy that effectively retains only the most informative visual tokens.
This significantly reduces inference latency with minimal impact on prediction accuracy for beam search.

Inspired by~\cite{omri2025token,wang2025internvl3}, we incorporate a visual token selection algorithm into the encoder output, specifically on the patch-level visual tokens produced by the ViT encoder, before they are fed into the BART encoder.
Given a query image with $V_{max}$ visual tokens, our method selects a subset $V_{sub}$ of the most relevant tokens for subsequent decoding. Specifically, during both training and inference, we apply K-means++~\cite{arthur2006k} clustering on the embeddings of the original $V_{max}$ visual tokens to form $V_{sub}$ cluster centers.
Within each cluster, only the token closest to the center is retained.
This procedure effectively groups similar patterns, highlights salient regions in the image, and filters out less informative signals.
We then adopt a distillation framework, where a frozen post-trained generative model serves as the teacher and the pruned model as the student. KL divergence is used to ensure consistency between their output distributions:
\begin{equation}
\mathcal{L}_{\mathrm{dis}}
=
\sum_{k=1}^{L}
\mathrm{KL}\bigl(
\pi_{\mathrm{ref}}(t_k \mid u,t_{<k})
\,\big\|\,
\pi_{\theta}(t_k \mid u, t_{<k})
\bigr),
\label{eq:9}
\end{equation}
where $\pi_{\mathrm{ref}}$ is the reference model and $\pi_{\mathrm{\theta}}$ is the policy model.

Finally, inspired by curriculum learning~\cite{wang2021survey}, we introduce a progressive pruning strategy, where the number of retained visual tokens $v^{(e)}$ gradually decreases as the training epoch $e$ increases, as formulated below:
\begin{equation}
v^{(e)} = \left\lfloor
V_{max} - \frac{e}{E}\bigl(V_{max} - V_{sub}\bigr)
\right\rfloor,
\quad e = 1,\dots,E,
\label{eq:10}
\end{equation}
where $E$ is the total number of training epochs and at the final epoch, the final retained visual token count is $v^{(E)}=V_{sub}$.

\subsection{End-to-End Inference}
\label{sec:3.4}
In this section, we describe the end-to-end inference of OneVision.
Given a query image and user-specific information, the model first applies the pruning strategy to compress the visual tokens from the encoder, reducing redundancy while preserving key semantic information. 
The decoder then autoregressively generates a sequence of discrete codes that represent potential matching items.

To produce a well-ranked candidate set, we use beam search. 
This method maintains multiple SID candidates during generation, selecting them based on the accumulated log-probability scores.
Starting with a $t_{[BOS]}$ token, the model predicts the top-$B$ most likely next tokens ($B$ being the beam size), and extends each token sequence. 
After scoring the extended sequences, only the top-$B$ with the highest cumulative scores are retained.
To prevent invalid SIDs, we apply constrained decoding using a Trie structure~\cite{de2020autoregressive,fredkin1960trie}. The Trie is built from the full set of valid SIDs before inference and restricts decoding to valid prefixes only.
This ensures that generated sequences are valid and improves retrieval reliability.

\begin{table*}[!htbp]
\caption{Comparison of Encoding Strategies and Retrieval Paradigms on CLICK and ORDER Benchmarks.
Ours (w/o P.) and Ours (w/ P.) denote the results without and with the incorporation of personalization signals and additional business information.}
\centering
\setlength{\tabcolsep}{4.5pt}
\footnotesize
\resizebox{\textwidth}{!}{%
\begin{tabular}{ll ccccc ccccc}
\toprule
\multirow{2}{*}{Methods} & \multirow{2}{*}{} 
& \multicolumn{5}{c}{CLICK (8.5K)} 
& \multicolumn{5}{c}{ORDER (8.5K)} \\
\cmidrule(lr){3-7} \cmidrule(lr){8-12}
& & HR@1 & HR@4 & MRR@4 & HR@10 & MRR@10 
  & HR@1 & HR@4 & MRR@4 & HR@10 & MRR@10 \\
\midrule
\multicolumn{2}{l}{\textbf{Traditional Retrieval}} \\
& DINOv3   & 49.68\% & 72.11\% & 54.78\% & 77.66\% & 55.63\% & 44.95\% & 70.92\% & 54.24\% & 76.80\% & 55.15\% \\
& OnlineMCA  & \underline{51.01\%} & \underline{74.82\%} & \underline{60.16\%} & \textbf{83.89\%} & \underline{61.37\%} & \underline{50.84\%} & \underline{74.92\%} & \underline{59.83\%} & \textbf{82.85\%} & \underline{60.51\%} \\
\midrule
\multicolumn{2}{l}{\textbf{Generative Retrieval}} \\
\multirow{4}{*}{RQ-Kmeans} 
& IRGen        & 38.29\% & 57.29\% & 46.19\% & 66.56\% & 47.72\% & 37.22\% & 56.26\% & 46.19\% & 66.13\% & 46.70\% \\
& GENIUS & 42.43\% & 62.50\% & 50.93\% & 70.72\% & 52.30\% & 41.19\% & 61.70\% & 49.74\% & 69.95\% & 51.15\% \\
& Ours (w/o P.)   & 45.30\% & 63.70\% & 52.95\% & 72.10\% & 54.36\% & 44.61\% & 63.05\% & 52.18\% & 72.04\% & 53.68\% \\
& Ours (w/ P.)   & 48.91\% & 68.67\% & 57.14\% & 77.41\% & 58.60\% & 48.05\% & 68.38\% & 56.37\% & 77.38\% & 57.88\% \\
\midrule
\multirow{4}{*}{VRQ} 
& IRGen        & 44.13\% & 64.72\% & 52.90\% & 71.04\% & 53.96\% & 41.71\% & 63.48\% & 50.86\% & 70.13\% & 52.01\% \\
& GENIUS & 45.22\% & 65.51\% & 54.00\% & 71.22\% & 54.98\% & 42.84\% & 64.05\% & 51.91\% & 70.43\% & 52.98\% \\
& Ours (w/o P.)   & 50.56\% & 73.51\% & 57.91\% & 78.50\% & 59.07\% & 48.57\% & 72.22\% & 58.20\% & 76.27\% & 57.47\% \\
& Ours (w/ P.)   & \textbf{51.46\%} & \textbf{77.28\%} & \textbf{61.77\%} & \underline{82.29\%} & \textbf{62.46\%} & \textbf{51.42\%} & \textbf{76.28\%} & \textbf{60.89\%} & \underline{80.23\%} & \textbf{61.71\%} \\
\bottomrule
\end{tabular}}
\label{tab: CDRM_online_dataset}
\end{table*}

\section{Experiment}

In this section, we present a comprehensive evaluation of OneVision using industrial-scale e-commerce datasets, along with rigorous online A/B testing. In addition, we conduct extensive ablation studies to verify the usability of OneVision for image retrieval and to further facilitate the real-world adoption of generative vision search systems in industrial e-commerce scenarios.

\subsection{Implementation Details}

\textbf{Datasets}. We construct training datasets from large-scale industrial e-commerce search logs to support the multi-stage training of OneVision. The corpus consists of three parts: (1) query images and product images, each denoted with its corresponding SIDs, providing discrete representations for both search queries and products; (2) item-to-item (i2i) associations, representing semantically similar product pairs; and (3) 4 months' data of user search logs from Kuaishou’s platform, covering item click and purchase records. Both the i2i and user-interaction data incorporate multi-view paired samples, enabling the model to build a more comprehensive and robust understanding of each product across diverse visual appearances. For evaluation, we built two test sets. The first is a real-world search interaction set, containing 8.5K click events and 8.5K purchase events sampled from 30 days of historical visual search traffic, designed to measure retrieval quality under actual user behaviours. The second is a semantic benchmark, comprising 67K query–item pairs (about 22K each from top, middle, and tail segments) collected via ANN recall. 
All datasets are additionally refined using ANN similarity scores and VLM verification to ensure reliable supervision and robust evaluation.
\\
\textbf{Evaluation Metrics}. The evaluation considers both recall and ranking performance, employing HitRate (HR) and Mean Reciprocal Ranking (MRR) as metrics, which are widely recognized in search and recommendation system research. All reported results represent averages computed over the entire set of experiments.
\\
\textbf{Baseline Methods}. To comprehensively evaluate OneVision's e-commerce image retrieval capabilities, we compare its performance with the state-of-the-art image feature extraction model (DINOv3) and two generative image retrieval approaches (IRGen and GENIUS)~\cite{simeoni2025dinov3,zhang2024irgen,kim2025genius}. For fairness, DINOv3 uses a ViT-B backbone; IRGen is configured with a 12-layer decoder (about 120M parameters); and GENIUS employs a BART backbone rather than T5. Additionally, we benchmark its product search effectiveness against the online multi-stage cascading architecture (referred to as onlineMCA) to further validate OneVision’s image-based search performance.
\\
\textbf{Details of the OneVision}. We employ a compact BART model with 6 encoder layers and 6 decoder layers, each with a hidden size of 768. For the VRQ discrete SID design, the total number of codebook layers is set to $L = 5$, including three RQ-VAE layers to capture hierarchical semantics and two additional residual OPQ layers for fine-grained residual refinement. The codebook size $K$ for each layer is configured as $(2048, 512, 256 \mid 256, 256)$. For offline evaluation, we use constrained beam search with a beam size of 10, while for online inference, the beam size is increased to 256 to balance generation quality and latency. 
The multi-view contrastive learning for RQ-VAE codebooks is performed following the SimCLR paradigm, using a batch size of 4096.
We adopt a CosineAnnealingLR scheduler with a maximum learning rate of $8\times10^{-5}$ for Pretraining, SFT, and DPO. The batch size is configured as 768 for both Pretraining and SFT, and 256 for DPO.

\begin{table}[htbp]
\caption{Ablation Study of VRQ with Different Encoding Designs and Additional Information Signals.}
\centering
\small
\setlength{\tabcolsep}{8pt} 
\renewcommand{\arraystretch}{1.2} 
\begin{adjustbox}{max width=\columnwidth}
\begin{tabular}{lcccc}
\toprule
\multirow{2}{*}{\textbf{Configuration}} &
\multicolumn{2}{c}{\textbf{CLICK (8.5K)}} &
\multicolumn{2}{c}{\textbf{ORDER (8.5K)}} \\
\cmidrule(lr){2-3}\cmidrule(lr){4-5}
& HR@10 & MRR@10 & HR@10 & MRR@10 \\
\midrule
onlineMCA                          & \textbf{83.89\%} & 61.37\% & \textbf{82.85\%} & 60.51\% \\
\textemdash w/o ranking                        & 81.45\% & 56.87\% & 80.14\% & 55.47\% \\
\hdashline
RQ–Kmeans       & 77.41\% & 58.60\% & 77.39\% & 57.88\% \\
RQ–OPQ (2/256)     & 80.72\% & 61.51\% & 79.82\% & 60.19\% \\
\hdashline
VRQ (Ours)                 & \underline{82.29\%} & \textbf{62.46\%} & \underline{80.23\%} & \textbf{61.71\%} \\
\textemdash w/o User-Behaviour SIDs                        & 81.91\% & 61.90\% & 79.71\% & 61.25\% \\
\textemdash w/o Convert$_{emb}$                  & 78.98\% & 59.72\% & 77.58\% & 58.02\% \\
\textemdash w/o Category$_{emb}$                           & 81.89\% & \underline{62.37\%} & 79.95\% & \underline{61.56\%} \\
\bottomrule
\end{tabular}
\end{adjustbox}
\label{tab:sft2-ablation}
\end{table}

\begin{table*}[!htbp]
\caption{Comparison of Retrieval Methods on the Semantic Benchmark across Top, Middle, and Long-tail segments.
Ours (Multi) and Ours (Single) represent the encoding of e-commerce multi-modal and single-modal visual features, respectively.}
\centering
\setlength{\tabcolsep}{5.5pt}
\renewcommand{\arraystretch}{1.1}
\scriptsize
\resizebox{0.9\textwidth}{!}{%
\begin{tabular}{p{2.6cm} ccc ccc ccc}
\toprule
\multirow{2}{*}{Methods}
& \multicolumn{3}{c}{60w Top}
& \multicolumn{3}{c}{60w Middle}
& \multicolumn{3}{c}{60w Long-tail} \\
\cmidrule(lr){2-4}\cmidrule(lr){5-7}\cmidrule(lr){8-10}
& HR@1 & HR@4 & MRR@4
& HR@1 & HR@4 & MRR@4
& HR@1 & HR@4 & MRR@4 \\
\midrule
\multicolumn{10}{l}{\textbf{Embedding-based Retrieval}} \\
DINOv3        & 79.12\% & 87.56\% & 84.14\% & 78.61\% & 88.86\% & 82.97\% & 77.46\% & 87.41\% & 82.10\% \\
Online Model$^{*}$    &  99.41\%   &  99.80\%   &  99.59\%   &  98.76\%   &  99.78\%   &  99.24\%   &  98.33\%   &  99.78\%   &  99.00\%   \\
Quantized ANN Score (QAS)     & 81.45\% & 90.34\% & 85.08\% & 72.77\% & 85.10\% & 77.74\% & 69.57\% & 83.54\% & 75.23\% \\
\midrule
\multicolumn{10}{l}{\textbf{Generative Retrieval}} \\
IRGen           &  \underline{88.49\%}   &  \underline{94.92\%}   &  \underline{91.28\%}   &  \underline{84.19\%}   &  \underline{92.65\%}   &  \underline{87.81\%}   &  \underline{80.42\%}   &  \underline{91.93\%}   &  \underline{85.32\%}   \\
GENIUS          & 86.63\% & 94.06\% & 89.84\% & 82.49\% & 91.78\% & 86.45\% & 78.88\% & 91.08\% & 84.09\% \\
VLM             &  85.77\%   &  91.80\%  &  88.41\%   &  77.01\%   &  88.51\%   &  81.92\%   &  71.24\%   &  85.94\%   &  77.29\%   \\
Ours (Multi)     & 68.61\% & 76.24\% & 72.15\% & 55.32\% & 67.56\% & 63.53\% & 49.12\% & 63.66\% & 58.32\% \\
Ours (Single)     & \textbf{89.29\%} & \textbf{96.53\%} & \textbf{93.30\%} & \textbf{85.19\%} & \textbf{93.70\%} & \textbf{89.55\%} & \textbf{81.89\%} & \textbf{92.15\%} & \textbf{86.09\%} \\
\bottomrule
\end{tabular}}
\label{tab:relevance-buckets}
\vspace{-0.5em}
\end{table*}

\subsection{Real-World Offline Evaluation}
\label{sec:4.2}
As shown in Table~\ref{tab: CDRM_online_dataset}, our method consistently improves retrieval performance across different visual encoding strategies and training paradigms on the real-world search interaction set, reaching and slightly surpassing OnlineMCA.
Concretely, the GRs encoded by RQ-Kmeans perform worse than systems using DINOv3 features with ANN retrieval. Replacing RQ–Kmeans with our VRQ results in consistent retrieval gains, with HR@1 improving by 8.64\% and HR@4 by 11.07\% on average over CLICK and ORDER test sets. 
This improvement stems from replacing unsupervised RQ–KMeans with a contrastively trained RQ-VAE to mitigate SID imbalance and cluster underutilization, while residual OPQ layers refine visual residuals for more uniform and semantically aligned SIDs.

Under the unified VRQ encoding, OneVision surpasses traditional generative models. Compared with IRGen and GENIUS, our multi-stage generative learning scheme improves HR@1/4 by 13.44\% on average and raises MRR@10 by 8.96\%. 
Furthermore, incorporating personalization signals and statistical business information brings an additional +3.22 HR points and +3.55 MRR points. Notably, the best VRQ (w/ P.) slightly surpasses OnlineMCA (e.g., CLICK HR@1 51.46\% vs 51.01\%, ORDER HR@1 51.42\% vs 50.84\%, and +1.15pt MRR@10), indicating that end-to-end generative retrieval can now rival and even exceed well-tuned ANN-based industrial systems, offering a new, unified paradigm for large-scale and highly dynamic e-commerce search applications.

Table~\ref{tab:sft2-ablation} presents the ablation study of OneVision. Removing the ranking stage from onlineMCA leads to a noticeable decline in retrieval quality, confirming its importance. Substituting RQ–Kmeans with RQ–OPQ raises HR and MRR improves HR and MRR by 4.04\% on average, showing that residual OPQ improves the stability of SIDs and enables a more fine-grained representation. 
The VRQ with multi-stage training reaches the highest overall scores while remaining comparable to the deployed online system. Adding statistical business characteristics (Convert) information brings an additional 2.98pt in HR and 3.22pt in MRR, which helps the model separate near-duplicate products more effectively. In contrast, user sequence modeling shows limited impact, likely because user preferences are sparse and relatively consistent, offering little additional signal for personalization. Category embeddings mainly mitigate category inconsistency and detection noise, lowering off-category retrievals and strengthening search robustness and user experience, while having a limited direct impact on HR or MRR.

\subsection{More Offline Evaluation }
\label{sec:4.3}

To rigorously validate the effectiveness of our proposed method, we construct an e-commerce semantic dataset and benchmark multiple retrieval approaches under the same conditions.

Table \ref{tab:relevance-buckets} compares embedding-based retrieval and generative retrieval across category tiers (Top / Middle / Long-tail). Quantizing the strong online embedding model for ANN retrieval causes a pronounced drop in accuracy (e.g., HR@1 decreases from 98.83\% to 74.60\% and MRR@4 from 99.28\% to 79.35\%), highlighting the sensitivity of embedding-based systems to aggressive quantization and the importance of robust encoding approaches.

Within generative retrieval, our image-centric GR model consistently outperforms other GR baselines across all category tiers. Compared with IRGen and GENIUS, it improves HR@1 by an average of 1.98\% on top categories and 2.81\% on long-tail categories, with similar gains observed for HR@4 and MRR@4. VLM shows promise in principle, but is not yet fully optimized for this setting; matching our results would demand substantially larger training data and computational resources. Notably, our design also surpasses multimodal encoding by a wide margin (e.g., Top HR@1 89.29\% vs. 68.61\%), underscoring that focusing on fine-grained visual features while preserving category consistency is both more effective and more practical for large-scale e-commerce retrieval.

Table \ref{tab:corr_ablation} summarizes the performance of multiple encoding approaches using several evaluation metrics, including ICO, QAS, and GR.
It can be clearly seen that methods focusing only on reconstruction quality, but ignoring semantic alignment, struggle when applied to generative decoding. For instance, although FSQ achieves extremely strong QAS results (HR@4 98.47\%, MRR@4 96.99\%), its performance collapses under GR (HR@4 36.56\%, MRR@4 23.32\%).

The table further highlights the benefits of contrastive training and residual OPQ refinement.
Among RQ-VAE variants, the Tiger baseline without contrastive objectives yields HR@4 79.60\% and MRR@4 68.80\%, while GENIUS, with contrastive learning, improves to 92.34\% and 87.15\%. Building on this, our VRQ strengthens the representation by adding OPQ-based residual decomposition, reaching HR@4 94.13\% and MRR@4 89.65\%. These results confirm that combining fine-grained residual encoding with contrastive learning produces more discriminative and generation-friendly SIDs.

\begin{figure}[!t]
    \centering
    \begin{subfigure}[t]{0.23\textwidth}
        \centering
        \includegraphics[width=\textwidth]{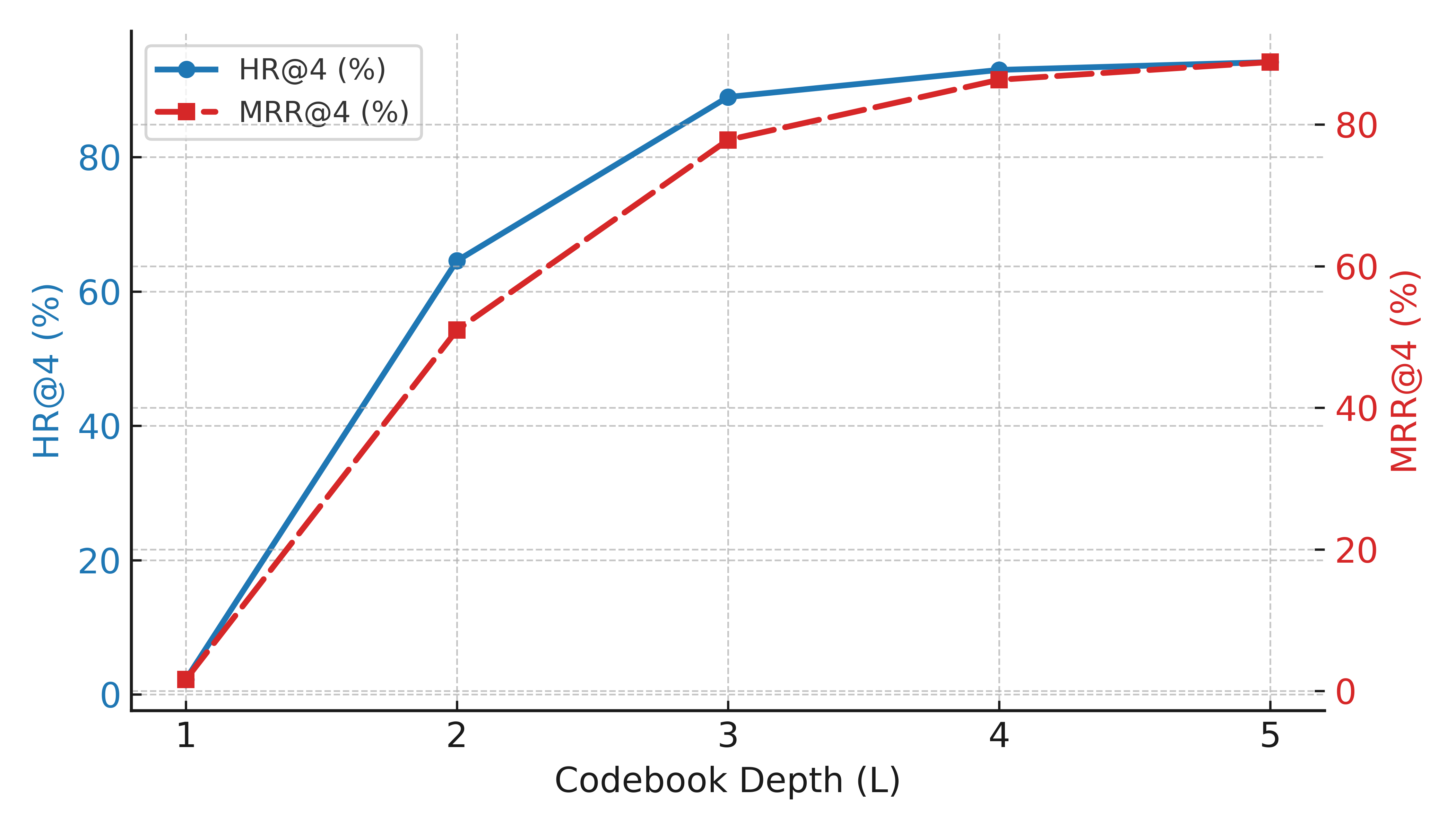}
        \caption{Results across depth $L$}
        \label{fig:subfig1_L}
    \end{subfigure}
    \hfill
    \begin{subfigure}[t]{0.23\textwidth}
        \centering
        \includegraphics[width=\textwidth]{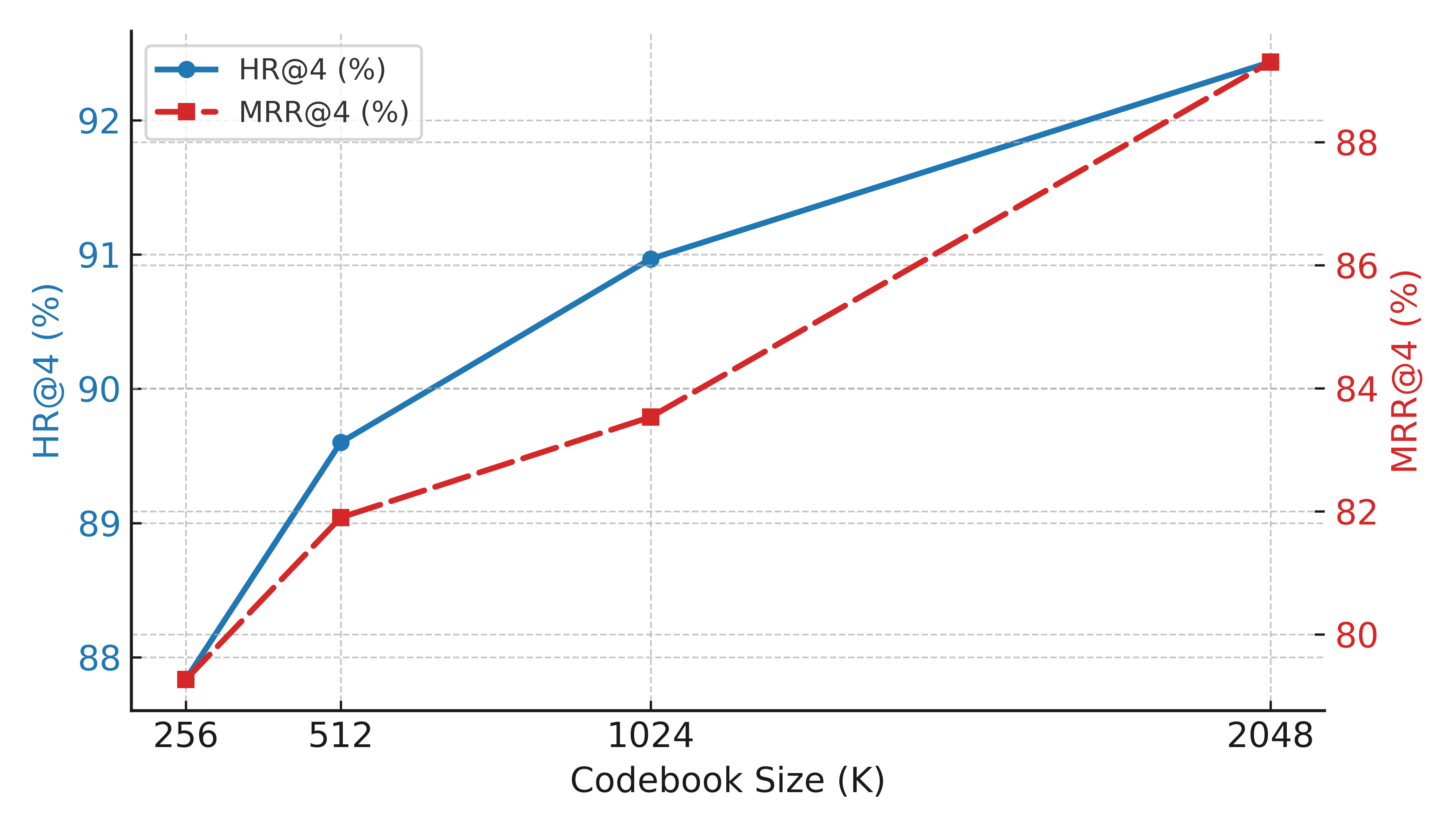}
        \caption{Results across size $K$}
        \label{fig:subfig1_K}
    \end{subfigure}
    \caption{Impact of Different Codebook Depth $L$ and Size $K$.}
    \label{fig:L and K ablation}
    \vspace{-1.2em}
\end{figure}

We also ablate the RQ-VAE codebook depth $L$ and size $K$ under constrained beam search. 
As shown in Figure~\ref{fig:L and K ablation}, increasing $L$ from $1$ to $4$ improves HR@4 from 2.21\% to 93.03\% and MRR@4 from 1.58\% to 86.33\%, with only marginal gains beyond $L=4$. 
Similarly, enlarging $K$ from $256$ to $2048$ raises HR@4 from 87.84\% to 92.50\% and MRR@4 from 80.27\% to 95.32\%. 
Overall, within a practical codebook space and constrained decoding, retrieval accuracy improves monotonically with increasing depth and size.

\begin{table}[!t]
\caption{Comparison of Encoding Methods on Independent Code Occupancy (ICO), Quantized ANN Score (QAS), and Generative Retrieval (GR).}
\centering
\setlength{\tabcolsep}{6pt}
\footnotesize
\renewcommand{\arraystretch}{1.2}
\begin{adjustbox}{max width=\columnwidth}
\begin{tabular}{l c cc cc}
\toprule
\multirow{2}{*}{Method} & \multirow{2}{*}{\makecell{
ICO
}}
& \multicolumn{2}{c}{
QAS
} & \multicolumn{2}{c}{
GR
} \\
\cmidrule(lr){3-4} \cmidrule(lr){5-6}
& & HR@4 & MRR@4 & HR@4 & MRR@4 \\
\midrule
RQ-Kmeans     &  4.84  &  81.76\%  &  74.66\%   &  89.98\%   &  85.42\%   \\
FSQ            &  2.17   &  \textbf{98.47\%}   &   \textbf{96.99\%}  &  36.56\%   &  23.32\%  \\
OPQ            &   4.83  &   84.70\%  &  79.05\%   &  86.07\%   &  78.83\%   \\
RQ-VAE (Tiger)          &  4.91   & 69.48\%    &  61.41\%   &  79.60\%   &  68.80\%   \\
RQ-VAE (GENIUS)        &  4.23   &  83.75\%   &  77.18\%   &  92.34\%   &  87.15\%   \\
RQ--OPQ (2/256)       &  3.92   &  85.15\%   &  78.23\%   &  \underline{93.28\%}   &  \underline{88.80\%}   \\
VRQ (Ours)      &  3.78   &  \underline{86.33\%}   &  \underline{79.35\%}   &  \textbf{94.13\%}   &  \textbf{89.65\%}   \\
\bottomrule
\end{tabular}
\end{adjustbox}
\label{tab:corr_ablation}
\vspace{-1.5em}
\end{table}

\begin{figure}[!t]
    \centering
    \begin{subfigure}[t]{0.23\textwidth}
        \centering
        \includegraphics[width=\textwidth]{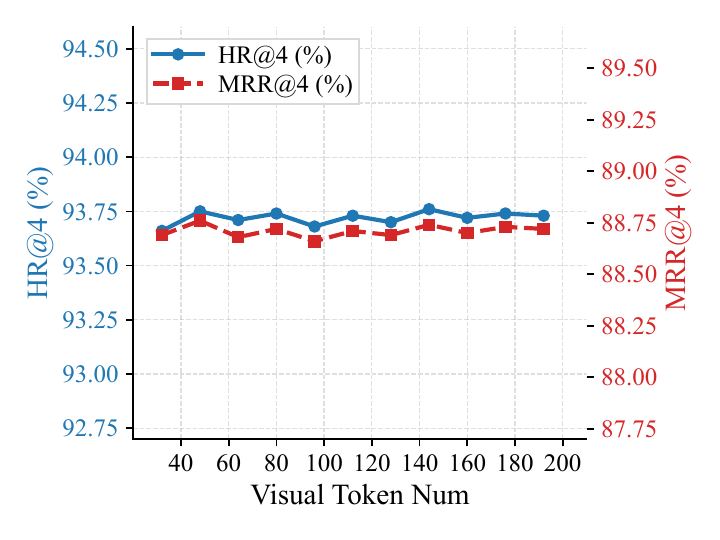}
        \caption{Accuracy vs. token count}
        \label{fig:subfig1_hr_mrr}
    \end{subfigure}
    \hfill
    \begin{subfigure}[t]{0.23\textwidth}
        \centering
        \includegraphics[width=\textwidth]{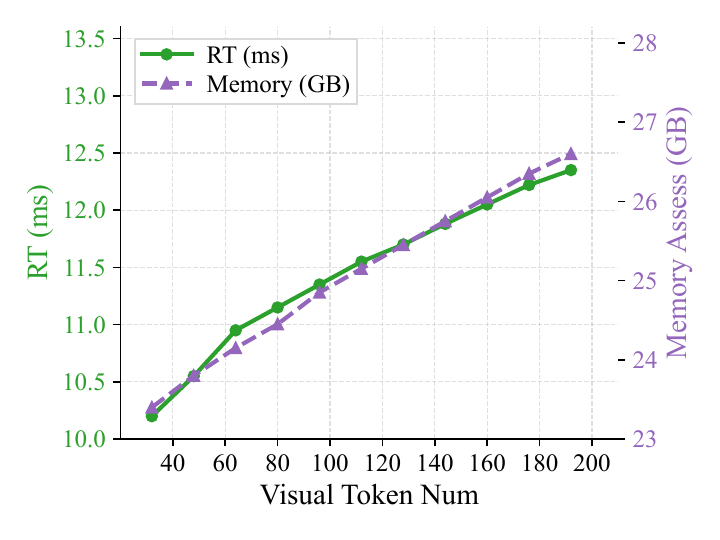}
        \caption{Efficiency vs. token count}
        \label{fig:subfig1_rt_mem}
    \end{subfigure}
    \caption{Impact of Visual Token Number on Retrieval Accuracy  (HR@4, MRR@4) and System Efficiency.}
    \label{fig:different visual token}
    \vspace{-1.2em}
\end{figure}

Finally, Figure~\ref{fig:different visual token} shows that with curriculum-based visual token pruning, the model can dynamically reduce the number of visual tokens while preserving retrieval accuracy. HR@4 and MRR@4 remain stable even when compressing from the full 197 tokens to only 33 tokens. This compression yields substantial efficiency gains: real-time inference latency (RT) drops by 17.41\% and memory usage decreases by 12.03\%. By reducing redundancy in the encoder’s visual representations and easing the burden of beam search decoding, this approach markedly lowers online inference cost while maintaining strong semantic understanding.

\subsection{Online A/B Testing}

To validate the effectiveness of OneVision in real-world Vision Search, we conduct rigorous online A/B tests on KuaiShou’s search platform, where the query is an image captured in real time or selected from the photo album. As indicated in Table~\ref{tab:online-metrics}, we establish one base group and two experimental groups : (1) OnlineMCA (baseline), (2) OnlineMCA w/o ranking that retains only recall + pre-ranking, and (3) the proposed OneVision. For OneVision, because identical product codes can still appear even after introducing statistical business signals, we maintain a dynamic item sequence for each code and sort items in descending order of

\vspace{-0.9em}
\begin{equation}
S_{\mathrm{conv}}
\;=\;
\omega_{1}\,\mathrm{Clicks}_{30\mathrm{d}}
\;+\;
\omega_{2}\,\mathrm{GMV}_{30\mathrm{d}}
\;+\;
\omega_{3}\,\mathrm{Orders}_{30\mathrm{d}} \,,
\label{eq:conv_score}
\end{equation}

\noindent
where $\omega_{1},\omega_{2},\omega_{3}$ are tunable weights that control the relative importance of recent clicks, GMV, and completed orders within a 30-day window.
This prioritizes items with stronger recent engagement and transaction value within a code cluster. Results show that multi-stage supervised training together with VRQ encoding yields consistent improvements over OnlineMCA: CTR +3.61\%, CVR +2.43\%, OPM +4.53\%, Click +3.43\%, and Order +5.89\%. In contrast, OnlineMCA w/o ranking exhibits marked declines (e.g., Order -12.65\%), underscoring the centrality of ranking in Photo Search and indirectly confirming that OneVision, augmented by its dynamic item sequence, already demonstrates strong ranking capability. Overall, these outcomes indicate that OneVision matches or surpasses the baseline OnlineMCA while simplifying the online stack without inducing seesaw effects, and it shows clear headroom for continued iteration in the era of large generative models.

\section{Further Analysis}

This section discusses three critical issues in the online deployment of our end-to-end generative vision search framework: analyzing where OneVision’s gains come from, highlighting its current limitations, and outlining potential avenues for future exploration.

\noindent\textbf{Sources of OneVision’s Online Gains.} The main sources of OneVision’s online gains can be summarized in two aspects. First, multi-view modeling enables the generative retrieval system to go beyond the rigid semantic constraints of the traditional ANN framework. By generating multiple SIDs that capture diverse viewpoints and appearances of the same product, it significantly improves the recall of identical items that conventional cascaded retrieval often misses when faced with large perspective variations. Second, a more unified and clear definition of similarity reduces the model’s reliance on low-level visual details and strengthens category-level semantic consistency, leading to more reliable retrieval results and greater conversion potential.

\begin{figure}
  \includegraphics[width=0.48\textwidth]{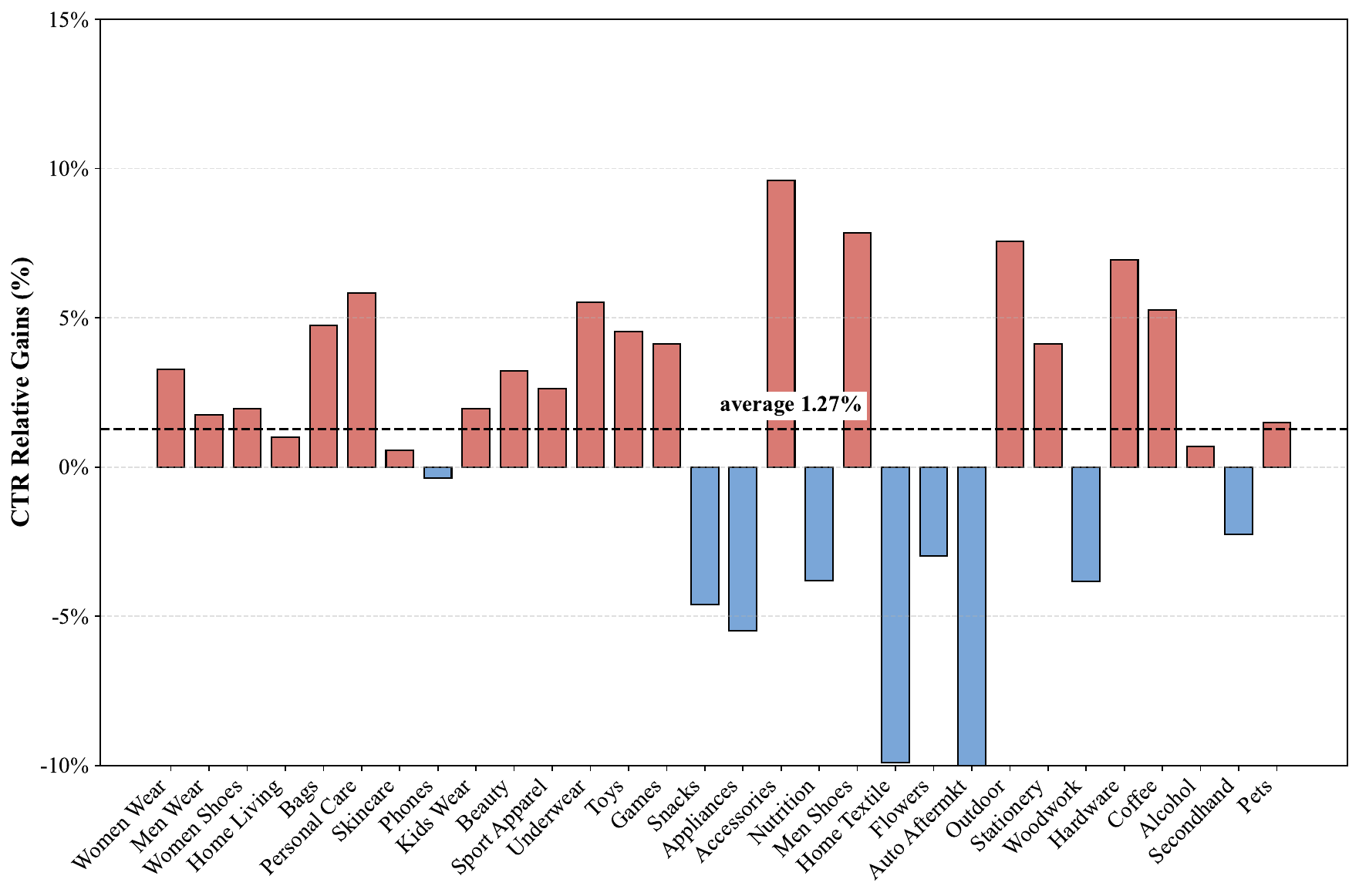}
  \centering
  \caption{Relative Item CTR Gains of Vision Search Across the Top 30 Industries.}
  \label{fig:cate_ctr}
\end{figure}

We further validate these two points through an in-depth analysis of the online A/B test results. With multi-view modeling, the overall Item CTR shows a clear +1.27\% relative gain across Top 30 categories. Notably, women wear improves by about +3.3\%, while bags increase by approximately +4.8\%. Meanwhile, unified similarity definition and improved category alignment lead to consistent PV CTR gains across category tiers: the top, middle, and long-tail segments increase by +2.58\%, +1.63\%, and +15.29\%, respectively. These results (see Figure~\ref{fig:cate_ctr} and Table~\ref{tab:click_rate_lift}) highlight OneVision's effectiveness in strengthening both identical-item retrieval and category-level consistency, ultimately driving measurable online CTR improvements and enhancing user experience.

\begin{table}[!t]
\caption{Online A/B Testing Results (All Improvements Statistically Significant, $p<0.05$)}
\centering
\setlength{\tabcolsep}{6pt}
\footnotesize
\renewcommand{\arraystretch}{1.2}
\begin{adjustbox}{max width=0.99\columnwidth}
\begin{tabular}{l c c c c c}
\toprule
Method & Item CTR & PV CTR & CVR & OPM & Order \\
\midrule
OnlineMCA w/o ranking & -5.01\% & -5.51\% & -10.73\% & -10.39\%  & -12.65\% \\
OneVision        & \textbf{+3.61\%} & \textbf{+3.43\%} & \textbf{+2.43\%} & \textbf{+4.53\%}  & \textbf{+5.89\%} \\
\bottomrule
\end{tabular}
\end{adjustbox}
\label{tab:online-metrics}
\end{table}

\begin{figure*}
  \includegraphics[width=1.0\textwidth]{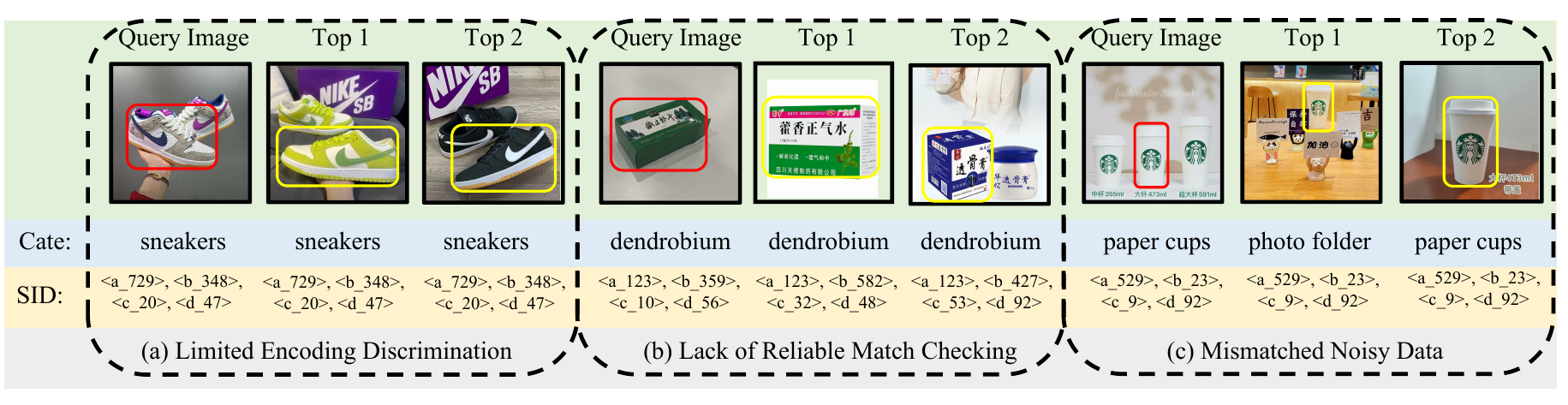}
  \centering
  \caption{An Overview of the Three Fundamental Challenges in Generative Image Retrieval, Including Limited Encoding Discrimination, Lack of Reliable Match Verification, and Mismatched Noise Data.}

  \label{fig:bad case}
\end{figure*}

\noindent\textbf{Limitations of Generative Image Retrieval.} Traditional generative image retrieval based on RQ-Kmeans encoding and one-stage training suffers from three key limitations. First, its coarse quantization lacks the fine-grained capacity to separate highly similar items, leading to insufficient discrimination among near-duplicate products. Second, it provides no principled mechanism to verify whether relevant candidates exist for a given query; naive reliance on raw decoder logits results in unstable and unreliable relevance estimation. Third, it fails to enforce category-level consistency and cannot effectively filter out candidates whose primary object is incorrectly recognized, leading to off-category retrievals (see Figure~\ref{fig:bad case}).

OneVision addresses these issues by combining RQ-VAE with OPQ to refine visual encoding and enlarge inter-item distinctions, incorporating an online Top-K similarity check to flag low-confidence retrievals, and injecting category information into the quantization process to preserve category alignment and reduce misclassification. Despite these advances, OneVision still lags behind ANN-based recall when queries are ambiguous, noisy, or belong to the long-tail with limited relevant training data, leading to lower recall and accuracy than dense ANN retrieval.

\noindent\textbf{Future Optimization Directions for OneVision.} Current generative image retrieval remains a cascaded design: encoding and generation are optimized separately, making it hard to achieve Pareto optimality in quality and efficiency. A key direction is joint training that unifies codebook learning and the generative model for better overall retrieval effectiveness. Meanwhile, VLMs are expected to serve as the backbone for future retrieval, offering deeper image–product understanding and better demand awareness. Exploring how to leverage large-model reasoning to strengthen VLM-based generative retrieval represents a promising avenue.

\begin{table}[!t]
\caption{Relative PV CTR Gains Across Top, Middle, and Long-Tail Categories.}
\centering
{\setlength{\tabcolsep}{12pt} 
\adjustbox{width=0.42\textwidth}{
\begin{tabular}{lccc}
\toprule
Method & Top & Middle & Long-tail \\
\midrule
\textit{OneVision} & +2.58\% & +1.63\% & +15.29\% \\
\bottomrule
\end{tabular}
}
}
\vspace{-1.2em}
\label{tab:click_rate_lift}
\end{table}

\section{Conclusion}
In summary, OneVision presents a unified generative framework that bridges retrieval and personalization in e-commerce vision search. By introducing vision-aligned residual quantization and multi-stage semantic alignment, it effectively handles multi-view discrepancies while preserving item distinctiveness. Experiments demonstrate notable gains in both offline and online settings, confirming that a semantic ID–driven generative paradigm can simplify the retrieval pipeline and enhance both accuracy and efficiency.

\bibliographystyle{unsrtnat} 
\balance
\bibliography{reference}

\end{document}